# A Characterization of the Dirichlet Distribution with Application to Learning Bayesian Networks

**Dan Geiger***
Computer Science Department
Technion, Haifa 32000, Israel
dang@cs.technion.ac.il

**David Heckerman**
Microsoft Research, Bldg 9S/1
Redmond WA, 98052-6399
heckerma@microsoft.com

## Abstract

We provide a new characterization of the Dirichlet distribution. This characterization implies that under assumptions made by several previous authors for learning belief networks, a Dirichlet prior on the parameters is inevitable.

## 1 Introduction

In recent years, several researchers have investigated Bayesian methods for learning belief networks [CH91, Bu91, SDLC93, HGC94]. These approaches all have the same basic components: a scoring metric and a search procedure. The scoring metric takes data and a network structure and returns a score reflecting the goodness-of-fit of the data to the structure. A search procedure generates networks for evaluation by the scoring metric. These approaches use the two components to identify a network structure or set of structures that can be used to predict future hypotheses or infer causal relationships.

The Bayesian approach can be described as follows. Suppose we have a domain of variables $\{u_1, \ldots, u_n\} = U$, and a set of cases $\{C_1, \ldots, C_m\} = D$ where each case is an instance of some or of all the variables in $U$. We sometimes refer to $D$ as a database. Let $(B_S, B_P)$ be a belief network, that is, $B_S$ is a directed acyclic graph, each node $i$ of $B_s$ is associated with a random variable $u_i$ and $B_P$ is a set of conditional distributions, $p(u_i|u_{i_1}, \ldots, u_{i_k})$, $1 \le i \le n$, where $u_{i_1}, \ldots, u_{i_k}$ are the variables corresponding to the parents of node $i$ in $B_S$. (For more details, consult [Pe88]). Let $B_S^h$ stand for the hypothesis that cases are drawn from a belief network having the structure $B_S$. Then a Bayesian measure of the goodness-of-fit of a belief network structure $B_S$ is $p(B_S^h|D, \xi)$ given by $p(B_S^h|D, \xi) = c \cdot p(B_S^h|\xi) p(D|B_S^h, \xi)$ where $c$ is a normalizing factor and $\xi$ is the current state of knowledge.

To compute $p(D|B_S^h, \xi)$ in closed form several assumptions were made. First, the database $D$ is a multinomial sample from some belief network $(B_S, B_P)$. Second, for each network structure the parameters associated with one node are independent of the parameters associated with other nodes (global independence [SL90]) and the parameters associated within a node given one instance of its parents are independent of the parameters of that node given other instances of its parent nodes (local independence [SL90]). Third, if a node has the same parents in two distinct networks then the distribution of the parameters associated with this node are identical in both networks (parameter modularity [HGC94]). Forth, each case is complete. Fifth, the distribution of the parameters associated with each node is Dirichlet.

The last two assumptions are made so as to create a conjugate sampling situation, namely, after data is seen the distributions of the parameters stay in the same family— the Dirichlet family. A relaxation of the assumption of complete cases was carried out by previous works (e.g., [SDLC93]). The contribution of this paper is a characterization of the Dirichlet distribution which enables one to show that the fifth assumption is implied from the first three assumptions and from one additional plausible assumption that if $B_1$ and $B_2$ are equivalent belief networks (i.e., they represent the same independence assumptions) then the events $B_1^h$ and $B_2^h$ are equivalent as well (hypothesis equivalence [HGC94]). We make this self-evident assumption explicit because it does not hold for causal networks where two edges with opposing directions correspond to distinct events.

Our contribution can be described using common statistical terminology as follows. We use this terminology because our result might be found applicable in other statistical uses of the Dirichlet distribution and because it falls under the broad area of characterizations of probability distribution functions. Suppose $s$ and $t$ are two discrete random variables having finite domains, $\{s_i\}_{i=1}^k$ and $\{t_j\}_{j=1}^n$, respectively. We wish to infer the joint probability $p(s, t)$ from a sample of pairs of values $(s_i, t_j)$ of $s$ and $t$. The standard Bayesian approach to this statistical inference problem is to associate with $p(s_i, t_j)$ a parameter $\theta_{ij}$ (often called the multinomial parameter), assign $\{\theta_{ij}|1 \le i \le k, 1 \le j \le n\}$ a prior joint pdf and compute the posterior joint pdf of $\{\theta_{ij}\}$ given the observed set of pairs of values. There are two closely-related variants to this approach which can be described as follows.

Let $\theta_{i\cdot} = \sum_{j=1}^{n} \theta_{ij}$ stand for the multinomial parameter associated with $p(s = s_i)$ and let $\theta_{j|i} = \theta_{ij} / \sum_j \theta_{ij}$ stand for the multinomial parameter associated with $p(t = t_j | s = s_i)$. Furthermore, let $\theta_{I\cdot} = \{\theta_{i\cdot}\}_{i=1}^{k-1}$ and $\theta_{J|i} = \{\theta_{j|i}\}_{j=1}^{n-1}$. We assume that $\{\theta_{I\cdot}, \theta_{J|1}, \ldots, \theta_{J|k}\}$ are mutually independent and that each has a prior pdf. Now according to Bayesian practice we compute the joint posterior appropriately. That is, we update the pdf for $\theta_{I\cdot}$ according to the counts of $s = s_i$ in the observed pairs and update the pdf of $\theta_{J|i}$ according to the counts of $t = t_j$ in all pairs in which $s = s_i$. In a symmetric fashion, let $\theta_{\cdot j} = \sum_{i=1}^{k} \theta_{ij}$, $\theta_{i|j} = \theta_{ij} / \sum_i \theta_{ij}$, $\theta_{\cdot J} = \{\theta_{\cdot j}\}_{j=1}^{n-1}$ and $\theta_{I|j} = \{\theta_{i|j}\}_{i=1}^{k-1}$. Now we assume that $\{\theta_{\cdot J}, \theta_{I|1}, \ldots, \theta_{I|n}\}$ are mutually independent and that each has a prior pdf and we compute the posterior pdf for $\theta_{\cdot J}$ according to the counts of $t = t_j$ and the posterior pdf of $\theta_{I|j}$ according to the counts of $s = s_i$ in all pairs in which $t = t_j$.

To make these techniques operational one must choose a specific prior pdf for the multinomial parameters. The standard choice of a pdf for $\{\theta_{ij}\}$ is a Dirichlet pdf usually for pragmatic reasons. When such a choice is made, it can be shown that $\{\theta_{I\cdot}, \theta_{J|1}, \ldots, \theta_{J|k}\}$ are indeed mutually independent and that each has a prior Dirichlet pdf. Similarly, $\{\theta_{\cdot J}, \theta_{I|1}, \ldots, \theta_{I|n}\}$ are mutually independent and each has a prior Dirichlet pdf.

The surprising result proved in this article is that if these independence assertions are assumed to hold, and under the assumption of (strictly) positive pdfs, then a prior Dirichlet pdf for $\{\theta_{ij}\}$ is the only possible choice. The assumption of strictly positive pdfs can possibly be dropped without affecting the conclusion but we have not carried out a proof of this claim. The implication of this result to learning Bayesian networks is discussed in Section 3. A preliminary account of analogous results for Gaussian networks is reported in Section 4.

## 2 Background and Technical Summary

The Dirichlet pdf is defined as follows. Let $\phi_1, \ldots, \phi_l$ be positive random variables that sum to 1. Then $\phi_1, \ldots, \phi_{l-1}$ have a Dirichlet pdf $f$ if

$$f(\phi_1, \ldots, \phi_{l-1}) = \frac{\Gamma(\sum_{i=1}^{l} \alpha_i)}{\prod_{i=1}^{l} \Gamma(\alpha_i)} \prod_{i=1}^{l} \phi_i^{\alpha_i - 1} \quad (1)$$

where $\phi_l = 1 - \sum_{i=1}^{l-1} \phi_i$ and $\alpha_i$ are positive constants (See, e.g., [De70, Wi62]).

We use the following conventions. Suppose $\{\theta_{ij}\}$, $1 \leq i \leq k$, $1 \leq j \leq n$, is a set of positive random variables that sum to 1. Let $\theta_{i\cdot}, \theta_{\cdot j}, \theta_{I\cdot}, \theta_{\cdot J}, \theta_{j|i}, \theta_{i|j}, \theta_{J|i}$, and $\theta_{I|j}$ be defined as in the introduction. Consequently, $\theta_{i\cdot}\theta_{j|i} = \theta_{\cdot j}\theta_{i|j}$ for every $i$ and $j$. Let $f_U$ be the joint pdf of $\{\theta_{ij}\}$, $f_I$ be the pdf of $\theta_{I\cdot}$, and $f_{J|i}$ be the pdf of $\theta_{J|i}$. Similarly, let $f_J$ be the pdf of $\theta_{\cdot J}$, and $f_{I|j}$ be the pdf of $\theta_{I|j}$. Finally, let $f_{IJ}$ be the joint pdf of $\theta_{I\cdot}, \theta_{J|1}, \ldots, \theta_{J|k}$ and $f_{JI}$ be the joint pdf of $\theta_{\cdot J}, \theta_{I|1}, \ldots, \theta_{I|n}$.

A Dirichlet pdf for $\{\theta_{ij}\}$ is given by

$$f_U(\{\theta_{ij}\}) = c \prod_{i=1}^{k} \prod_{j=1}^{n} \theta_{ij}^{\alpha_{ij} - 1} \quad (2)$$

where $\theta_{kn} = 1 - \sum_A \theta_{ij}$, $A = \{(i,j) | 1 \leq i, j \leq n, i \neq k \text{ or } j \neq n\}$, $c$ is the normalization constant and $\alpha_{ij}$ are positive constants.

We observe that $f_U$ and $f_{IJ}$ are related through a change of variables. Since both $\{\theta_{i\cdot}\}_{i=1}^{k}$ and $\{\theta_{j|i}\}_{j=1}^{n}$ are defined in terms of $\{\theta_{ij}\}$ and since $\theta_{ij} = \theta_{i\cdot}\theta_{j|i}$, there exists a one-to-one and onto correspondence between $\{\theta_{ij}\}$ and $\{\theta_{i\cdot}\} \cup \{\theta_{j|i}\}$. The Jacobian $J_{k,n}$ of this transformation is given by

$$J_{kn} = \prod_{i=1}^{k} \theta_{i\cdot}^{n-1} \quad (3)$$

[HGC95].

The following lemma provides a known property of the Dirichlet distribution. A slightly weaker version is stated in [DL93] (Lemma 7.2).

**Lemma 1** *Let $\{\theta_{ij}\}$, $1 \leq i \leq k$, $1 \leq j \leq n$, where $k$ and $n$ are integers greater than 1, be a set of positive random variables having a Dirichlet distribution. Then, $f_I(\theta_{I\cdot})$ is Dirichlet, $f_{J|i}(\theta_{J|i})$ is Dirichlet for every $i$, $1 \leq i \leq k$, and $\{\theta_{I\cdot}, \theta_{J|1}, \ldots, \theta_{J|k}\}$ are mutually independent.*

**Proof:** Set $\theta_{ij} = \theta_{i\cdot}\theta_{j|i}$ in Eq. 2, multiply by $J_{kn}$, and regroup terms. □

The main claim of this article is that, under the assumption of a positive pdf for $\{\theta_{ij}\}$, the converse holds as well. More specifically, we prove the following theorem.

**Theorem 2** *Let $\{\theta_{ij}\}$, $1 \leq i \leq k$, $1 \leq j \leq n$, $\sum_{ij} \theta_{ij} = 1$, where $k$ and $n$ are integers greater than 1, be positive random variables having a positive pdf $f_U(\{\theta_{ij}\})$. If $\{\theta_{I\cdot}, \theta_{J|1}, \ldots, \theta_{J|k}\}$ are mutually independent and $\{\theta_{\cdot J}, \theta_{I|1}, \ldots, \theta_{I|n}\}$ are mutually independent, then $f_U(\{\theta_{ij}\})$ is Dirichlet.*

Recall that $f_U$ can be written both in terms of $f_{IJ}$ and in terms of $f_{JI}$ by a change of variables and using the Jacobian given by Equation 3. Since both representations must be equal, and using the independence assumptions made by Theorem 2 to factor $f_{IJ}$ and $f_{JI}$, we get the equality,

$$\left(\prod_{j=1}^{n} \theta_{\cdot j}^{k-1}\right)^{-1} f_J(\theta_{\cdot J}) \prod_{j=1}^{n} f_{I|j}(\theta_{I|j}) = \quad (4)$$

$$\left(\prod_{i=1}^{k} \theta_{i\cdot}^{n-1}\right)^{-1} f_I(\theta_{I\cdot}) \prod_{i=1}^{k} f_{J|i}(\theta_{J|i})$$

This equality, which is in fact a functional equation, summarizes the independence assumptions stated in Theorem 2.

Methods for solving functional equations such as Eq. 4, that is, finding all functions that satisfy them under different regularity assumptions, are discussed in [Ac66]. We use the following technique. First, we show that any positive solution to Eq. 4 must be differentiable in any order (Aczél, 66, Section 4.2.2, "Deduction of differentiability from integrability"). Then we take repeated derivatives of Eq. 4 and obtain a differential equation the solution of which after appropriate specialization is the general solution of Eq. 4 (Aczél, 66, Section 4.2, "Reduction to differential equations"). The proof is given is the appendix.

Note that when $n = k = 2$ and by renaming of variable and function names, Eq. 4 can be written as follows:

$$f_0(y)g_1(z)g_2(w) = g_0(x)f_1\left(\frac{yz}{x}\right)f_2\left(\frac{y(1-z)}{1-x}\right) \quad (5)$$

where

$$x = yz + (1-y)w$$

and where $y$, $z$ and $w$ replace $\theta_{\cdot j=1}$, $\theta_{i=1|j=1}$, $\theta_{i=1|j=2}$, respectively.

## 3  Implications For Learning

We now explain how our characterization applies to learning belief networks. We concentrate on belief networks for two discrete variables $s$ and $t$ whose joint distribution is $p(s,t)$. The $n$-variate case is discussed in [HGC95]. There are three possible belief networks with two nodes. The network that contains no edge between its two nodes $s$ and $t$, a network $s \to t$ and the network $t \to s$. The first network $B_0$ corresponds to the assertion that $s$ and $t$ are independent while the second network $B_1$ and the third one $B_2$ assert that $s$ and $t$ are dependent. The last two belief networks are equivalent, $B_1$ represents the factorization $p(s,t) = p(s)p(t|s)$ and $B_2$ represents the factorization $p(s,t) = p(t)p(s|t)$.

We shall first examine the two complete networks $B_1$ and $B_2$. We assume that if two networks $B_1$ and $B_2$ are equivalent (as is the case in our example) then the corresponding events $B_1^h$ and $B_2^h$ are equivalent (hypothesis equivalence [HGC94]). Recalling the notations introduced in the introduction, we have that $\theta_{i\cdot} = \sum_{j=1}^{n} \theta_{ij}$ stand for the multinomial parameters associated with $p(s = s_i)$ and $\theta_{j|i} = \theta_{ij}/\sum_j \theta_{ij}$ stand for the multinomial parameters associated with $p(t = t_j | s = s_i)$. Thus,

$$f_{IJ}(\theta_{I\cdot}, \theta_{J|1}, \ldots, \theta_{J|k}|B_1^h) = f_{IJ}(\theta_{I\cdot}, \theta_{J|1}, \ldots, \theta_{J|k}|B_2^h)$$

$$f_{JI}(\theta_{\cdot J}, \theta_{I|1}, \ldots, \theta_{I|k}|B_2^h) = f_{JI}(\theta_{\cdot J}, \theta_{I|1}, \ldots, \theta_{I|k}|B_1^h)$$

Due to these equalities and using local and global independence to factor $f_{IJ}$ and $f_{JI}$, we immediately obtain Equation 4 (dropping the conditioning events is valid because $B_1^h$ and $B_2^h$ are equivalent). Thus for the two *complete networks* the only possible prior on their parameters is, according to Theorem 2, the Dirichlet distribution.

Note that we only use three assumptions: a multinomial sampling situation, local and global independence, and hypothesis equivalence. Implicitly, since we condition on $B_i^h$, is the assumption that each complete structure has a positive probability to be manifested.

The prior for any non-complete network follows from the assumption of parameter modularity which says that the pdf associated with a node under the assumption that a specific network generates the data is the same as the pdf of the parameters of that node given another network generates the data provided that the set of parents is identical in the two networks. In our two-variables network, for example, the parameters $\theta_{i\cdot}$ which are associated with node $s$ have the same pdf when conditioned on $B_1$ and when conditioned on $B_0$ because in both networks $s$ has the same set of parents (the empty set) and similarly for node $t$. That is,

$$f_i(\theta_{i\cdot}|B_1^h, \xi) = f_i(\theta_{i\cdot}|B_0^h, \xi)$$

$$f_j(\theta_{\cdot j}|B_2^h, \xi) = f_j(\theta_{\cdot j}|B_0^h, \xi)$$

These equalities imply that the prior for the parameters of $B_0$ is Dirichlet as well. Thus, parameter modularity is the assumption that extends our result from complete to non-complete networks.

This result of the inevitable choice of a Dirichlet prior for two-variables networks is easily generalized to the $n$-variate case by induction and without the need to solve any additional functional equations. The inductive proof uses the fact that a cluster of variables each having a Dirichlet distribution is distributed Dirichlet as well. For details consult [HGC95].

Recall that the exponents of $\theta_{ij}$ of a Dirichlet distribution can be written as $N\alpha_{ij} - 1$ where $N$ is the "equivalent sample size" (the size of an imaginary database of complete cases–the prior sample–upon which the prior Dirichlet is based) and $\alpha_{ij}$ is the expectation of $\theta_{ij}$. The equivalent sample size reflects the confidence of the user and $\alpha_{ij}$ represents the relative frequency of the pair $(i,j)$ in the prior sample. A joint Dirichlet prior is therefore quite restricting because it allows only one equivalent sample size for the entire domain. That is, there is no way to express different confidence levels regarding the parameters of different parts of the network. Thus the practical ramification of our characterization is that the commonly-made global and local independence assumption is inappropriate whenever a single equivalent sample size is not sufficient to describe prior knowledge. Such a situation occurs, for example, if knowledge about $\theta_{I\cdot}$ is more precise than knowledge about $\theta_{J|i}$.

One possibility for overcoming this limitation of the Dirichlet prior is to replace the notion of a single equivalent sample size with *equivalent database*. Namely, we ask a user to imagine that she was initially completely ignorant about a domain, having an uninformative prior with equivalent sample size(s) close to the lower bound. Then, we ask the user to specify a database $D_e$ that would produce a posterior density that reflects her current state of knowledge. This database may contain incomplete cases. Then, to score a real database $D$, we score the database $D_e \cup D$, using the uninformative prior and a learning algorithm that handles missing data. This way of specifying a prior yields a mixture of Dirichlet distributions which, according to our result, cannot satisfy the local and global independence assumption.

## 4 Discussion

The independence assumptions made by Theorem 2 can be divided into two parts: $\{\theta_{J|1}, \ldots, \theta_{J|k}\}$ are mutually independent and $\{\theta_{I|1}, \ldots, \theta_{I|n}\}$ are mutually independent (local independence) and $\theta_I$ is independent of $\{\theta_{J|1}, \ldots, \theta_{J|k}\}$ and $\theta_J$ is independent of $\{\theta_{I|1}, \ldots, \theta_{I|n}\}$ (global independence). A natural question to ask is whether global independence alone implies a joint Dirichlet pdf for $\{\theta_{ij}\}$.

This question is particularly interesting in light of the analysis of decomposable graphical models given by [DL93]. Dawid and Lauritzen term a pdf that satisfies global independence *a strong hyper-Markov law* and show the importance of such laws in the analysis of decomposable graphical models. We now show that the class of strong hyper-Markov laws is larger than the Dirichlet class.

When $n = k = 2$, and using the notations of Equation 5 the new functional equation can be written as follows:

$$f_0(y)g(z,w) = g_0(x)f(\frac{yz}{x}, \frac{y(1-z)}{1-x}) \qquad (6)$$

where $x = yz + (1-y)w$. Note that Eq. 5 is obtained from this equation by setting $g(z,w) = g_1(z)g_2(w)$ and $f(t_1, t_2) = f_1(t_1)f_2(t_2)$. These equalities correspond to local independence.

Let $f_U$ be a joint pdf of $\{\theta_{ij}\}$ given by

$$f_U(\{\theta_{ij}\}) = K \left[\prod_{i=1}^{2}\prod_{j=1}^{2} \theta_{ij}^{\alpha_{ij}-1}\right] H(\frac{\theta_{11}\theta_{22}}{\theta_{12}\theta_{21}}) \qquad (7)$$

where $K$ is the normalization constant, $\alpha_{ij}$ are positive constants and $H$ is an arbitrary positive integrable function. That this pdf satisfies global independence can be easily verified. It can in fact be shown, by solving Eq. 6, that every positive strong Hyper Markov law can be written in this form (when $n = 2$ and $k = 2$). This solution includes the Dirichlet family as a proper subclass.

Since $H$ is a single function that does not depend on a particular network, one can conclude that if local parameter independence is assumed to hold in *one network*, then $f_U$ must still be Dirichlet and therefore, due to Lemma 1, local parameter independence must hold for *all networks*. We have so far proved this claim for two-variables networks but we believe it holds for the $n$-variate case as well.

As a final comment, we should mention that a functional equation which restricts the possible prior distributions for the parameters of Bayesian networks can be formulated for other sampling situations not necessarily for the multinomial sampling which was assumed in our discussion so far. As another example, consider a two-continuous-variables domain $\{x_1, x_2\}$ having a bivariate-normal distribution. Constructing a prior for the parameters of such Gaussian networks and performing the prior-to-posterior analysis was carried out in [GH94, HG95]. Let $\{m_1, v_1, m_{2|1}, b_{12}, v_{2|1}\}$ and $\{m_2, v_2, m_{1|2}, b_{21}, v_{1|2}\}$ denote the parameters for the network structures $x_1 \rightarrow x_2$ and $x_1 \leftarrow x_2$, respectively. That is, $m_1$ is the mean of $x_1$ and $v_1$ is the variance for $x_1$. Collectively, these are the parameters associated with node $x_1$ in the first network. The parameters associated with node $x_2$ are the conditional mean $m_{2|1}$, the regression coefficient $b_{12}$ of $x_2$ given $x_1$ and the conditional variance $v_{2|1}$. Now assuming global parameter independence and hypothesis equivalence and using the Jacobian given in [HG95] yields the functional equation

$$f_1(m_1, v_1) \, f_{2|1}(m_{2|1}, b_{12}, v_{2|1}) = \frac{v_1^2 v_{2|1}^3}{v_2^2 v_{1|2}^3} \cdot$$
$$f_2(m_2, v_2) \, f_{1|2}(m_{1|2}, b_{21}, v_{1|2}) \qquad (8)$$

where $f_1$, $f_{2|1}$, $f_2$, and $f_{1|2}$ are arbitrary density functions, and where

$$v_2 = v_{2|1} + v_1 b_{12}^2 \qquad b_{21} = \frac{b_{12} v_1}{v_2} \qquad v_{1|2} = \frac{v_{2|1} v_1}{v_2}$$

$$m_2 = m_{2|1} + b_{12} m_1 \qquad m_{1|2} = m_1 + b_{21} m_2$$

These relationship are well known from path analysis and can be derived from Eq. 4 in [HG95].

We have solved this functional equation and found that the only integrable solutions are such that $f_1(v_1)$ is an inverse gamma distribution, that is, $1/v_1$ has a gamma distribution, $f_1(m_1|v_1)$ is a normal distribution, and similarly for $f_2(m_2, v_2)$. The conditional distribution $f_{2|1}(b_{12}, v_{2|1})$ has an interesting form. An inverse gamma distribution for $v_{2|1}$ times a Normal distribution for $b_{12}$ times an arbitrary function $H(b_{12}/v_{2|1})$. The arbitrary function is not surprising since the functional equation only encodes global independence and so the solution depends on an arbitrary function just as for multinomial sampling (Equation 7).

The natural question is now what does local independence mean for Gaussian networks. Because the subjective variance of $b_{12}$ actually depends on $v_{2|1}$, we cannot assume that $b_{12}$ and $v_{2|1}$ are independent. The answer is that local independence for Gaussian networks means that the *standardized* regression coefficient $b_{12}$ is independent of the conditional variance $v_{2|1}$ at each

200    Geiger and Heckermannode. When adding this assumption, which fully parallels the discrete case, we get that $H$ is the exponential function and therefore $f_{2|1}(b_{12}|v_{2|1})$ is a normal distribution and $f_{2|1}(v_{2|1})$ is an inverse Gamma distribution.

Consequently, it can further be shown that a bivariate normal-Wishart distribution is the only possible prior on the joint space parameters (i.e., the inverse covariance matrix and the vector of means) if we assume global parameter independence, local parameter independence for *one* network and hypothesis equivalence. Indeed this was the prior chosen by [GH94]. An analogous result holds for the $n$-variate case as well.

## Acknowledgment

We thank J. D. Aczél and M. Ungarish for valuable comments, S. Altschuler and L. Wu for their help with Lemma 3, M. Israeli for his help with Lemma 4 and G. Cooper who helped us define the notion of an equivalent database.

## References

[Ac66] J. Aczél, *Lectures on Functional Equations and Their Applications*, Academic Press, New York, 1966.

[Bu91] W. Buntine, Theory refinement on Bayesian networks, Proceedings of Seventh Conference on Uncertainty in Artificial Intelligence, Los Angeles, CA, Morgan Kaufmann, July, 1991, 52-60.

[CH91] G. Cooper and E. Herskovits, A Bayesian method for the induction of probabilistic networks from data, Section on Medical Informatics, Stanford University, January, 1991, Technical Report, SMI-91-1. Also in, Proceedings of Seventh Conference on Uncertainty in Artificial Intelligence, Los Angeles, CA, Morgan Kaufmann, July, 1991, 52-60. Final version in Machine Learning, 1992, 9:309-347.

[DL93] P. Dawid and S. Lauritzen, Hyper Markov laws in statistical analysis of decomposable graphical models, *Annals of Statistics*, 21:1272-1317, 1993.

[De70] M. DeGroot, *Optimal Statistical Decisions*, McGraw-Hill, New York.

[GH94] D. Geiger and D. Heckerman, Learning Gaussian networks, Proceedings of Tenth Conference on Uncertainty in Artificial Intelligence, Seattle, WA, Morgan Kaufmann, July 1994, 235-243.

[GH95] D. Geiger and D. Heckerman, A characterization of the Dirichlet distribution through global and local independence, Computer Science Department, Technical report 9506, February 1995. A preliminary report appears as Microsoft Research Report, TR-94-16.

[HG95] D. Heckerman and D. Geiger, Learning Bayesian networks: A unification for discrete and Gaussian domains. In this proceedings.

[HGC94] D. Heckerman, D. Geiger, and D. Chickering, Learning Bayesian networks: The combination of knowledge and statistical data, Proceedings of Tenth Conference on Uncertainty in Artificial Intelligence, Seattle, WA, Morgan Kaufmann, July 1994, 293-301.

[HGC95] D. Heckerman, D. Geiger, and D. Chickering, Learning Bayesian networks, Machine Learning, 1995, to appear.

[Pe88] J. Pearl, *Probabilistic Reasoning in Intelligent Systems: Networks of Plausible Inference*, 1988, Morgan Kaufmann, San Mateo, CA.

[SL90] D. Spiegelhalter and S. Lauritzen, Sequential updating of conditional probabilities on directed graphical structures, *Networks*, 20, 579-605, 1990.

[SDLC93] D. Spiegelhalter, A. Dawid, S. Lauritzen, and R. Cowell, Bayesian analysis in expert systems, *Statistical Science*, 8, 219-282, 1993.

[Wi62] S. Wilks, *Mathematical Statistics*, Wiley and Sons, New York.

## Appendix: Proof[1]

### Differeniability from Integrability

By renaming of variable and function names, and by absorbing the Jacobians into the new function definitions, Eq. 4 can be written as follows:

$$f_0(y_1, \ldots, y_{n-1}) \prod_{j=1}^{n} g_j(z_{1,j}, \ldots, z_{k-1,j}) =$$

$$g_0(x_1, \ldots, x_{k-1}) \prod_{i=1}^{k} f_i\left(\frac{z_{i1} y_1}{x_i}, \ldots, \frac{z_{i,n-1} y_{n-1}}{x_i}\right) \quad (9)$$

where

$$x_i = \sum_{j=1}^{n} z_{ij} y_j, \quad 1 \leq i \leq k-1 \quad (10)$$

$$z_{kj} = 1 - \sum_{i=1}^{k-1} z_{ij}, \quad 1 \leq j \leq n$$

and where

$$y_n = 1 - \sum_{j=1}^{n-1} y_j, \quad x_k = 1 - \sum_{i=1}^{k-1} x_i \quad (11)$$

Note that the free variables in Eq. 9 are $y_1, \ldots, y_{n-1}$ ($y_j$ replaces $\theta_{\cdot j}$) and $z_{ij}$, $1 \leq i \leq k-1$, $1 \leq j \leq n$ ($z_{ij}$ replaces $\theta_{i|j}$). All other variables which appear in Eq. 9 are defined by Eqs. 10 and 11.

---
[1]This proof first appeared in [GH95].

Furthermore, we may consider $x_1 \ldots, x_k$ ($x_i$ replaces $\theta_i$.) and $w_{ij} = \frac{z_{ij} y_j}{x_i}$, $1 \leq i \leq k$, $1 \leq j \leq n-1$ ($w_{ij}$ replaces $\theta_{j|i}$) to be free variables and rewrite Eq. 9 in term of these variables, namely,

$$g_0(x_1, \ldots, x_{k-1}) \prod_{i=1}^{k} f_i(w_{i,1}, \ldots, w_{i,n-1}) =$$
$$f_0(y_1, \ldots, y_{n-1}) \prod_{j=1}^{n} g_j(\frac{w_{1,j} x_1}{y_j}, \ldots, \frac{w_{k-1,j} x_{k-1}}{y_j}) \quad (12)$$

where

$$y_j = \sum_{i=1}^{k} w_{ij} x_i, \quad 1 \leq j \leq n-1 \quad (13)$$

$$w_{in} = 1 - \sum_{j=1}^{n-1} w_{ij}, \quad 1 \leq i \leq k$$

and where $x_k$ and $y_n$ are defined by Eq. 11. This symmetric representation of Eq. 9 will be used in the derivation of its solution.

We assume that all functions mentioned in Eq. 9 originated from pdfs and thus are (Lebesgue) integrable in their domain. We shall now show that this assumption implies that each set of functions that solves Eq. 9 consists of functions for which any finite-order partial derivative exists for every point in their domain. The importance of this technical claim is that in order to find all positive integrable functions that satisfy Eq. 9, it is permissible to take any derivative at any point in the domain because it exists.

By setting $z_{ij} = 1/k$, for all $i$ and $j$, in Equation 9 we get that $f_0(y_1, \ldots, y_{n-1})$ is proportional to $\prod_{i=1}^{k} f_i(y_1, \ldots, y_{n-1})$. Similarly, by setting $w_{ij} = 1/n$ in Eq. 12, $g_0(x_1, \ldots, x_{k-1})$ is proportional to $\prod_{j=1}^{n} g_j(x_1, \ldots, x_{k-1})$. Thus if we prove that each $g_j$, $j = 1, \ldots, n$, has any-order derivative, then so does $g_0$. Furthermore, any property that we prove about $g_j$, $j = 1, \ldots, n$, holds true for $f_i$, $i = 1, \ldots, k$, due to the symmetric representation of Eq. 9 given by Eq. 12.

Since all functions are positive, we can take the logarithm of Eq. 9. Since all functions are integrable and positive then so are their logarithms. Let now $j_0$ be an index such that $1 \leq j_0 \leq n$. We take a logarithm of Eq. 9 and integrate the resulting equation wrt [2] all variables except for the variables $z_{i,j_0}$, $1 \leq i < k$. Consequently, we obtain,

$$\hat{g}_{j_0}(z_{1,j_0}, \ldots, z_{k-1,j_0}) = M + \int_{D_{j_1}} \cdots \int_{D_{j_n}} \int_{D_y} [\hat{g}_0(x_1,$$
$$\ldots, x_{k-1}) + \sum_{i=1}^{k} \hat{f}_i(\frac{z_{i1} y_1}{x_i}, \ldots, \frac{z_{i,n-1} y_{n-1}}{x_i})] dZ_{j_0} dY \quad (14)$$

where $\hat{h}(x)$ stands for $\ln h(x)$, $M$ is a constant, $Y = (y_1, \ldots, y_{n-1})$, $Z_{j_0}$ is a vector containing all variables $z_{ij}$ except those where $j = j_0$, $D_j$ is the domain of $g_j$, and $D_y$ the domain of $f_0$.

Since, the right hand-side of Eq. 14 is integrable, it follows that $g_{j_0}$ is continuous for every $1 \leq j_0 \leq n$. Hence, $g_0$ is continuous as well. Thus, due to the symmetric functional equation (Eq. 12), $f_i$ are also continuous functions. Having now continuous functions on the right hand-side of Eq. 14, it follows that $g_{j_0}$ has a first derivative wrt each of its arguments. Thus, due to Eq. 12, each $f_i$ also has a first derivative wrt each of its arguments. Consequently, by Eq. 14, it follows that $g_{j_0}$ has a second derivative wrt each of its arguments. Repeating this argument yields the desired conclusion that all positive integrable functions that solve Equation 9 have any partial derivative at any point in their domain.[3]

## The Binary Solution

We shall now find all positive integrable solutions of Eq. 9 when $k = n = 2$. This derivation is different from the general derivation which is given in the following sections, however, the basic method of repeatedly differentiating the functional equation and subsequently solving the resulting differential equations is similar.

When $n = k = 2$, the functional equation can be written as follows:

$$f_0(y) g_1(z) g_2(w) = g_0(x) f_1(\frac{yz}{x}) f_2(\frac{y(1-z)}{1-x}) \quad (15)$$

where

$$x = yz + (1-y)w \quad (16)$$

Let

$$\hat{f}'_i(t) = \frac{d}{dt} \ln f_i(t) \quad (17)$$

and

$$\hat{g}'_i(t) = \frac{d}{dt} \ln g_i(t) \quad (18)$$

Taking the logarithm and then a derivative once wrt $y$, once wrt $z$ and once wrt $w$ of Eq. 15 yields the following three equations,

$$\hat{f}'_0(y) - (z-w) \hat{g}'_0(x) = \frac{zw}{x^2} \hat{f}'_1(\frac{yz}{x}) + \frac{(1-z)(1-w)}{(1-x)^2} \hat{f}'_2(\frac{y(1-z)}{1-x}) \quad (19)$$

$$\hat{g}'_1(z) - y \hat{g}'_0(x) = \frac{yw(1-y)}{x^2} \hat{f}'_1(\frac{yz}{x}) - \frac{(1-w)(1-y)y}{(1-x)^2} \hat{f}'_2(\frac{y(1-z)}{1-x}) \quad (20)$$

$$\hat{g}'_2(w) - (1-y) \hat{g}'_0(x) = -\frac{yz(1-y)}{x^2} \hat{f}'_1(\frac{yz}{x}) + \frac{y(1-z)(1-y)}{(1-x)^2} \hat{f}'_2(\frac{y(1-z)}{1-x}) \quad (21)$$

---

[2] with respect to

[3] Note that, by definition, a pdf does not include a delta function. Otherwise it is called a generalized pdf (gpdf, [De70]). An integral of a gpdf need not be continuous.

Solving $\hat{f}'_1(\frac{yz}{x})$ and $\hat{f}'_2(\frac{y(1-z)}{1-x})$ from Eqs. 20 and 21 yields,

$$\frac{y(1-y)(w-z)}{x^2}\hat{f}'_1(\frac{yz}{x}) = -(1-y)(1-w)\hat{g}'_0(x)$$
$$+(1-z)\hat{g}'_1(z) - y(1-z)\hat{g}'_0(x) + (1-w)\hat{g}'_2(w) \quad (22)$$

$$\frac{y(1-y)(w-z)}{(1-x)^2}\hat{f}'_2(\frac{y(1-z)}{1-x}) = z\hat{g}'_1(z)$$
$$- yz\hat{g}'_0(x) + w\hat{g}'_2(w) - (1-y)w\hat{g}'_0(x) \quad (23)$$

Now plugging Eqs. 22 and 23 into Eq. 19 and collecting all the terms involving $\hat{g}'_0(x)$, $\hat{g}'_1(z)$, $\hat{g}'_2(w)$ and $\hat{f}'_0(y)$, yields,

$$h(y,z,w)\hat{g}'_0(x) = z(1-z)\hat{g}'_1(z) + w(1-w)\hat{g}'_2(w)$$
$$- y(1-y)(w-z)\hat{f}'_0(y) \quad (24)$$

where

$$h(y,z,w) = y(1-y)(w-z)^2 + yz(1-z)$$
$$+ (1-y)(1-w)w$$

Taking a derivative wrt $z$ of Eq. 24 yields,

$$h_z(y,z,w)\hat{g}'_0(x) + yh(y,z,w)\hat{g}''_0(x) = (1-2z)\hat{g}'_1(z)$$
$$+ z(1-z)\hat{g}''_1(z) + y(1-y)\hat{f}'_0(y) \quad (25)$$

where $h_z$ is the partial derivative of $h$ wrt $z$, given by,

$$h_z(y,z,w) = -2y(1-y)(w-z) + y(1-2z)$$

Similarly, taking a derivative wrt $w$ of Eq. 24 yields,

$$h_w(y,z,w)\hat{g}'_0(x) + (1-y)h(y,z,w)\hat{g}''_0(x) =$$
$$(1-2w)\hat{g}'_2(w) + w(1-w)\hat{g}''_2(w) - y(1-y)\hat{f}'_0(y) \quad (26)$$

where $h_w$ is the partial derivative of $h$ wrt $w$, given by,

$$h_w(y,z,w)\hat{g}'_0(x) = 2y(1-y)(w-z) + (1-y)(1-2w)$$

Eqs. 25 and 26 together with

$$(1-y)h_z(y,z,w) - yh_w(y,z,w) \equiv 0$$

yield

$$(1-2w)\hat{g}'_2(w) + w(1-w)\hat{g}''_2(w) = (1-y)\hat{f}'_0(y)$$
$$+ \frac{1-y}{y}[(1-2z)\hat{g}'_1(z) + z(1-z)\hat{g}''_1(z)]$$
$$(27)$$

Since $w$ does not appear in the right hand side of this equation, we get,

$$(1-2w)\hat{g}'_2(w) + w(1-w)\hat{g}''_2(w) = c_1 \quad (28)$$

where $c_1$ is an arbitrary constant. Eq. 28 is a first order linear differential equation the general solution of which is given by,

$$\hat{g}'_2(w) = \frac{b}{w(1-w)} - \frac{c_1}{2}\frac{1-2w}{w(1-w)}$$

where $b$ is an arbitrary constant and $\frac{b}{w(1-w)}$ is the homogeneous solution. Thus,

$$\hat{g}'_2(w) = \frac{\alpha}{w} - \frac{\beta}{1-w}$$

where $\alpha$ and $\beta$ are arbitrary constants defined by $\alpha = b - \frac{c_1}{2}$ and $\beta = -(b + \frac{3c_1}{2})$. Hence, using $\int \alpha/w\,dw = \ln w^\alpha$, we get $g_2(w) = cw^\alpha(1-w)^\beta$ where $c$ is a third arbitrary constant.

From Eq. 27 we also get,

$$(1-2z)\hat{g}'_1(z) + z(1-z)\hat{g}''_1(z) = \frac{c_1 y}{1-y} + y\hat{f}'_0(y)$$

Hence both sides are equal to a constant, say $c_2$. Consequently,

$$\hat{f}'_0(y) = \frac{c_2}{y} - \frac{c_1}{1-y}$$

and

$$\hat{g}'_1(z) = \frac{\alpha'}{z} - \frac{\beta'}{1-z}$$

Consequently, $f_0(y)$, $g_1(z)$ and $g_2(w)$ all have the Dirichlet functional form and each function depends on three constants. Due to the symmetric representation of Eq. 9 given by Eq. 12, we conclude that $g_0$, $f_1$, and $f_2$ are Dirichlet as well.

## Preliminary Lemmas

We now provide several lemmas that are needed for the derivation of the general solution of Eq. 9.

**Lemma 3** *The general solution of the following partial differential equation for $f(x_1,\ldots,x_n)$,*

$$f + x_i f_{x_i} + x_j f_{x_j} = 0 \quad (29)$$

*in the domain $(0,\infty)^m$, is given by,*

$$f(x_1,\ldots,x_n) = \frac{1}{x_i}h(\frac{x_i}{x_j}, x_1,\ldots,x_{i-1},x_{i+1},\ldots,$$
$$x_{j-1},x_{j+1},\ldots,x_n) \quad (30)$$

*or, equivalently, by*

$$f(x_1,\ldots,x_n) = \frac{1}{x_j}g(\frac{x_i}{x_j}, x_1,\ldots,x_{i-1},x_{i+1},\ldots,$$
$$x_{j-1},x_{j+1},\ldots,x_n) \quad (31)$$

*where $h$ and $g$ are arbitrary differentiable functions having $n-1$ arguments.*

**Proof:** Let $s = x_i$ and $t = \frac{x_i}{x_j}$. Thus, $f_{x_i} = f_s + \frac{t}{s}f_t$, $f_{x_j} = -\frac{t^2}{s}f_t$. Hence, after a change of variables, the differential equation becomes

$$f + sf_s = 0$$

and therefore,

$$f = \frac{1}{s}h(t, x_1,\ldots,x_{i-1},x_{i+1},\ldots,x_{j-1},x_{j+1},\ldots,x_n)$$

By changing the roles of $x_i$ and $x_j$ in this derivation, we get the other form of $f$. □

**Lemma 4** *The general solution of the following partial differential equation for $f(x_1, \ldots, x_n)$,*

$$f_{x_i} - f_{x_j} = \frac{\alpha}{x_i} + \frac{\beta}{x_j} \quad (32)$$

*is given by*

$$f(x_1, \ldots, x_n) = \alpha \ln x_i - \beta \ln x_j + h(x_i + x_j, x_1, \ldots, x_{i-1}, x_{i+1}, \ldots, x_{j-1}, x_{j+1}, \ldots, x_n) \quad (33)$$

*where $h$ is an arbitrary differentiable function having $n - 1$ arguments.*

**Proof:** Let $s = x_i + x_j$ and $t = x_i - x_j$. Thus, $f_{x_i} = f_s + f_t$, $f_{x_j} = f_s - f_t$. Hence, after a change of variables, the differential equation becomes

$$f_t = \frac{\alpha}{s+t} + \frac{\beta}{s-t}$$

Integrating wrt $t$ and changing back to the original variables yields the desired solution. □

**Lemma 5** *Let $f(x_1, \ldots, x_n)$ be a twice-differentiable function. If for all $1 \leq i < j \leq n$,*

$$f(x_1, \ldots, x_n) = a_i \ln x_i + a_j \ln x_j + f_{ij}(x_i + x_j, x_1, \ldots, x_{i-1}, x_{i+1}, \ldots, x_{j-1}, x_{j+1}, \ldots, x_n)$$

*where $f_{ij}$ are arbitrary twice differentiable functions having $n - 1$ arguments, then*

$$f(x_1, \ldots, x_n) = g(\sum_{i=1}^n x_i) + \sum_{i=1}^n a_i \ln x_i \quad (34)$$

*where $g$ is an arbitrary twice-differentiable function.*

**Proof:** We shall prove the following stronger claim. For every $2 \leq l \leq n$, and for every permutation of the indices of $x_1, \ldots, x_n$,

$$f(x_1, \ldots, x_n) = h_l(\sum_{i=1}^l x_i, x_{l+1}, \ldots, x_n) + \sum_{i=1}^l a_i \ln x_i \quad (35)$$

where $h_l$ is an arbitrary twice differentiable function. The function $h_l$ depends on the permutation, although this fact is not reflected in our notation. The base case $l = 2$ is assumed by the lemma and the case $l = n$ is needed to be proven.

By the induction hypothesis we assume Eq. 35 and for the permutation

$$(1, \ldots, n) \to (l, 1, \ldots, l-1, l+1, \ldots, n)$$

we also assume (by the induction hypothesis),

$$f(x_1, \ldots, x_n) = g_l(x_l, x_{l+1} + \sum_{i=1}^{l-1} x_i, x_{l+2}, \ldots, x_n)$$
$$+ \sum_{i=1}^{l-1} b_i \ln x_i + b_{l+1} \ln x_{l+1} \quad (36)$$

Let $x = \sum_{i=1}^{l-1} x_i$, $c_i = b_i - a_i$ and $\overline{x} = (x_{l+2}, \ldots, x_n)$. From Eqs. 35 and 36 we get,

$$h_l(x_l + x, x_{l+1}, \overline{x}) = g_l(x_l, x_{l+1} + x, \overline{x}) +$$
$$\sum_{i=1}^{l-1} c_i \ln x_i - a_l \ln x_l + b_{l+1} \ln x_{l+1} \quad (37)$$

Set $x_i = 1/2(l-1)$, $i = 1, \ldots, l-1$. Thus, $x = 1/2$ and Eq. 37 yields,

$$h_l(x_l + 1/2, x_{l+1}, \overline{x}) = g_l(x_l, x_{l+1} + 1/2, \overline{x}) +$$
$$\sum_{i=1}^{l-1} c_i \ln(1/2(l-1)) - a_l \ln x_l + b_{l+1} \ln x_{l+1} \quad (38)$$

Plugging Eq. 38 into Eq. 37 and letting

$$\tilde{g}_l(x_l, x_{l+1} + x, \overline{x}) \equiv g_l(x_l, x_{l+1} + x, \overline{x}) - a_l \ln x_l \quad (39)$$

yields,

$$\tilde{g}_l(x_l + x - 1/2, x_{l+1} + 1/2, \overline{x}) = \tilde{g}_l(x_l, x_{l+1} + x, \overline{x}) +$$
$$\sum_{i=1}^{l-1} c_i \ln x_i - \sum_{i=1}^{l-1} c_i \ln(2(l-1)) \quad (40)$$

By taking a derivative wrt $x_j$, $1 \leq j \leq l-1$ of Eq. 40 we get,

$$\tilde{g}_l(x_l + x - 1/2, x_{l+1} + 1/2, \overline{x})_1 =$$
$$c_j/x_j + \tilde{g}_l(x_l, x_{l+1} + x, \overline{x})_2 \quad (41)$$

where the indices 1 and 2 indicate the argument of $\tilde{g}_l$ wrt which a derivative is taken. Similarly by taking the derivatives wrt $x_l$ we get,

$$\tilde{g}_l(x_l + x - 1/2, x_{l+1} + 1/2, \overline{x})_1 = \tilde{g}_l(x_l, x_{l+1} + x, \overline{x})_1 \quad (42)$$

Consequently,

$$\tilde{g}_l(x_l, x_{l+1} + x, \overline{x})_1 - \tilde{g}_l(x_l, x_{l+1} + x, \overline{x})_2 = c_j/x_j \quad (43)$$

for $j = 1, \ldots, l-1$.

we now show that $c_j = 0$. If $l > 2$, then set $j = j_1$ and $j = j_2$, $1 \leq j_1 < j_2 \leq l-1$, in Eq. 43 and subtract the two equations. Consequently, $c_{j_1}/x_{j_1} = c_{j_2}/x_{j_2}$ and therefore $c_{j_1} = c_{j_2} = 0$. If $l = 2$, then, $x = x_1$ and Eq. 43 becomes

$$\tilde{g}_l(x_2, x_3 + x_1, \overline{x})_1 - \tilde{g}_l(x_2, x_3 + x_1, \overline{x})_2 = c_1/x_1 \quad (44)$$

Let $u = x_1 + x_3$, $w = x_1 - x_3$ and rewrite the last equation,

$$\tilde{g}_l(x_2, u, \overline{x})_1 - \tilde{g}_l(x_2, u, \overline{x})_2 = \frac{2c_1}{u+w} \quad (45)$$

Since the left hand side is not a function of $w$ we have $c_1 = 0$.

Now let $s = x_l + (x + x_{l+1})$, $t = x_l - (x + x_{l+1})$ and rewrite the differential equation (Eq. 43) by changing variables to $s, t$ and $\overline{x}$. Since $c_j = 0$, we get,

$$\frac{\partial}{\partial t}[\tilde{g}_l(s, t, \overline{x})] = 0 \quad (46)$$

Thus, $\tilde{g}_l(s,t,\overline{x}) = i(s,\overline{x})$ where $i$ is a function of just $s$ and $\overline{x}$. Consequently, by switching back to the original variables, we get,

$$f(x_1,\ldots,x_n) = \sum_{i=1}^{l+1} a_i \ln x_i + i(\sum_{i=1}^{l+1} x_i, x_{l+2},\ldots,x_n) \quad (47)$$

Since this equation can be derived for any permutation of the indices of $x_i$, the induction is completed. $\square$

## 5  The General Solution

We now solve Eq. 9 for any $n$ and $k$. First we assume both $n$ and $k$ are strictly greater than 2.

We use the following notations:

$$g_l(t_1,\ldots,t_{k-1})_i = \frac{\partial}{\partial t_i} \ln g_l(t_1,\ldots,t_{k-1}) \quad (48)$$

$$g_l(t_1,\ldots,t_{k-1})_{ij} = \frac{\partial}{\partial t_i}\frac{\partial}{\partial t_j} \ln g_l(t_1,\ldots,t_{k-1})$$

$$f_l(t_1,\ldots,t_{n-1})_i = \frac{\partial}{\partial t_i} \ln f_l(t_1,\ldots,t_{n-1})$$

$$f_l(t_1,\ldots,t_{n-1})_{ij} = \frac{\partial}{\partial t_i}\frac{\partial}{\partial t_j} \ln f_l(t_1,\ldots,t_{n-1})$$

Also we use the following notations:

$$X = (x_1,\ldots,x_{k-1}), \qquad Z_j = (z_{1,j},\ldots,z_{k-1,j}), \quad (49)$$
$$Y = (y_1,\ldots,y_{n-1}), \qquad W_i = (\frac{z_{i,1}y_1}{x_i},\ldots,\frac{z_{i,n-1}y_{n-1}}{x_i})$$

For example, $g_j(Z_j)$ stands for $g_j(z_{1,j},\ldots,z_{k-1,j})$.

By taking the logarithm and then a derivative wrt $z_{ij}$ ($1 \leq i \leq k-1, 1 \leq j \leq n-1$) of Eq. 9, we get,

$$g_j(Z_j)_i = y_j \left[\sum_{l=1}^{n-1} f_i(W_i)_l \left[-\frac{z_{il}y_l}{x_i^2}\right] + \frac{1}{x_i} f_i(W_i)_j\right] + \quad (50)$$
$$y_j \left[\sum_{l=1}^{n-1} f_k(W_k)_l \left[\frac{z_{kl}y_l}{x_k^2}\right] - \frac{1}{x_k} f_k(W_k)_j\right] + y_j g_0(X)_i$$

By setting $i = i_1$ and $i = i_2$, $1 \leq i_1 < i_2 \leq k-1$ ($k \geq 3$) in Eq. 50, subtracting the resulting two equations, and dividing by $y_j$, we get,

$$\frac{1}{y_j}[g_j(Z_j)_{i_1} - g_j(Z_j)_{i_2}] = [g_0(X)_{i_1} - g_0(X)_{i_2}] +$$
$$\sum_{l=1}^{n-1} \left[f_{i_2}(W_{i_2})_l \left[\frac{z_{i_2 l}y_l}{x_{i_2}^2}\right] - f_{i_1}(W_{i_1})_l \left[\frac{z_{i_1 l}y_l}{x_{i_1}^2}\right]\right]$$
$$+ \frac{1}{x_{i_1}} f_{i_1}(W_{i_1})_j - \frac{1}{x_{i_2}} f_{i_2}(W_{i_2})_j \quad (51)$$

Taking now the logarithm and then a derivative wrt $z_{in}$ ($1 \leq i \leq k-1$) of Eq. 9 yields,

$$g_n(Z_n)_i = y_n g_0(X)_i + y_n \left[\sum_{l=1}^{n-1} f_i(W_i)_l \left[-\frac{z_{il}y_l}{x_i^2}\right]\right] +$$

$$y_n \left[\sum_{l=1}^{n-1} f_k(W_k)_l \left[\frac{z_{kl}y_l}{x_k^2}\right]\right] \quad (52)$$

Similarly, by setting $i = i_1$ and $i = i_2$, $1 \leq i_1 < i_2 \leq k-1$ in Eq. 52, subtracting the resulting two equations, and dividing by $y_n$, we get,

$$\frac{1}{y_n}[g_n(Z_n)_{i_1} - g_n(Z_n)_{i_2}] = g_0(X)_{i_1} - g_0(X)_{i_2} + \quad (53)$$
$$\sum_{l=1}^{n-1} \left[f_{i_2}(W_{i_2})_l \left[\frac{z_{i_2 l}y_l}{x_{i_2}^2}\right] - f_{i_1}(W_{i_1})_l \left[\frac{z_{i_1 l}y_l}{x_{i_1}^2}\right]\right]$$

Subtracting Eq. 53 from Eq. 51 and setting $j = j_1$, yields,

$$\frac{g_{j_1}(Z_{j_1})_{i_1} - g_{j_1}(Z_{j_1})_{i_2}}{y_{j_1}} - \frac{g_n(Z_n)_{i_1} - g_n(Z_n)_{i_2}}{y_n} =$$
$$\frac{f_{i_1}(W_{i_1})_{j_1}}{x_{i_1}} - \frac{f_{i_2}(W_{i_2})_{j_1}}{x_{i_2}} \quad (54)$$

where $1 \leq i_1 < i_2 \leq k-1, 1 \leq j_1 \leq n-1$.

Now we take a derivative wrt $z_{i_1 j_1}$ of Eq. 54 and obtain,

$$\frac{1}{y_{j_1}}[g_{j_1}(Z_{j_1})_{i_1 i_1} - g_{j_1}(Z_{j_1})_{i_2 i_1}] = -\frac{y_{j_1}}{x_{i_1}^2} f_{i_1}(W_{i_1})_{j_1} \quad (55)$$
$$+ \frac{y_{j_1}}{x_{i_1}} \sum_{l=1}^{n-1} f_{i_1}(W_{i_1})_{j_1 l} \left[-\frac{z_{i_1 l}y_l}{x_{i_1}^2}\right] + \frac{y_{j_1}}{x_{i_1}} f_{i_1}(W_{i_1})_{j_1 j_1}$$

Similarly, we take a derivative wrt $z_{i_1 n}$ of Eq. 54 and obtain,

$$-\frac{1}{y_n}[g_n(Z_n)_{i_1 i_1} - g_n(Z_n)_{i_2 i_1}] = -\frac{y_n}{x_{i_1}^2} f_{i_1}(W_{i_1})_{j_1} +$$
$$\frac{y_n}{x_{i_1}} \sum_{l=1}^{n-1} f_{i_1}(W_{i_1})_{j_1 l} \left[-\frac{z_{i_1 l}y_l}{x_{i_1}^2}\right] \quad (56)$$

Eqs. 55 and 56 yield,

$$\frac{1}{y_{j_1}^2}[g_{j_1}(Z_{j_1})_{i_1 i_1} - g_{j_1}(Z_{j_1})_{i_2 i_1}] + \frac{1}{y_n^2}[g_n(Z_n)_{i_1 i_1}$$
$$- g_n(Z_n)_{i_2 i_1}] = \frac{1}{x_{i_1}^2} f_{i_1}(W_{i_1})_{j_1 j_1} \quad (57)$$

Now, we take a derivative wrt $z_{i_1 j_2}$ of Eq. 54 where $1 \leq j_2 \leq n-1$, $j_2 \neq j_1$ ($n \geq 3$), and obtain,

$$0 = -\frac{y_{j_2}}{x_{i_1}^2} f_{i_1}(W_{i_1})_{j_1} + \frac{y_{j_2}}{x_{i_1}} \sum_{l=1}^{n-1} f_{i_1}(W_{i_1})_{j_1 l} \left[-\frac{z_{i_1 l}y_l}{x_{i_1}^2}\right]$$
$$+ \frac{y_{j_2}}{x_{i_1}^2} f_{i_1}(W_{i_1})_{j_1 j_2} \quad (58)$$

Eqs. 56 and 58 yield ($j_1 \neq j_2$),

$$\frac{1}{y_n^2}[g_n(Z_n)_{i_1 i_1} - g_n(Z_n)_{i_2 i_1}] = \frac{1}{x_{i_1}^2} f_{i_1}(W_{i_1})_{j_1 j_2} \quad (59)$$

Putting Eqs. 57 and 59 into Eq. 58 and recalling (from Eq. 10) that

$$z_{i_1 n} y_n = x_{i_1} - \sum_{l=1}^{n-1} z_{i_1 l} y_l$$

we get,

$$\frac{1}{x_{i_1}} f_{i_1}(W_{i_1})_{j_1} = -\frac{z_{i_1 j_1}}{y_{j_1}} [g_{j_1}(Z_{j_1})_{i_1 i_1} - g_{j_1}(Z_{j_1})_{i_2 i_1}]$$
$$+ \frac{z_{i_1 n}}{y_n} [g_n(Z_n)_{i_1 i_1} - g_n(Z_n)_{i_2 i_1}] \quad (60)$$

Similarly, we derive an analogue to Eq. 55 by taking a derivative wrt $z_{i_2 j_1}$ (instead of wrt $z_{i_1 j_1}$) of Eq. 54, follow the same steps up to Eq. 60, and get,

$$\frac{1}{x_{i_2}} f_{i_2}(W_{i_2})_{j_1} = -\frac{z_{i_2 j_1}}{y_{j_1}} [g_{j_1}(Z_{j_1})_{i_1 i_2} - g_{j_1}(Z_{j_1})_{i_2 i_2}]$$
$$+ \frac{z_{i_2 n}}{y_n} [g_n(Z_n)_{i_1 i_2} - g_n(Z_n)_{i_2 i_2}] \quad (61)$$

Plugging Eqs. 60 and 61 into Eq. 54 and collecting all terms involving $y_n$ in one side and all terms not involving $y_n$ on the other side implies that each side is equal to a constant, say $c$, namely,

$$\frac{1}{y_j}[g_j(Z_j)_{i_1} - g_j(Z_j)_{i_2}] + \frac{z_{i_1 j}}{y_j}[g_j(Z_j)_{i_1 i_1} - g_j(Z_j)_{i_2 i_1}]$$
$$+ \frac{z_{i_2 j}}{y_j}[g_j(Z_j)_{i_1 i_2} - g_j(Z_j)_{i_2 i_2}] = c \quad (62)$$

where $1 \leq j \leq n$.

This equation holds for every value of $y_j$ and therefore $c = 0$. Thus we obtain,

$$[g_j(Z_j)_{i_1} - g_j(Z_j)_{i_2}] + z_{i_1 j}[g_j(Z_j)_{i_1 i_1} - g_j(Z_j)_{i_2 i_1}]$$
$$+ z_{i_2 j}[g_j(Z_j)_{i_1 i_2} - g_j(Z_j)_{i_2 i_2}] = 0 \quad (63)$$

Let $h(Z_j) = g_j(Z_j)_{i_1} - g_j(Z_j)_{i_2}$. Thus Eq. 63 can be written as follows,

$$h + z_{i_1 j} \frac{\partial h}{\partial z_{i_1 j}} + z_{i_2 j} \frac{\partial h}{\partial z_{i_2 j}} = 0 \quad (64)$$

Lemma 3 provides the general solution for $h$ and thus,

$$h(Z_j) = g_j(Z_j)_{i_1} - g_j(Z_j)_{i_2} = \frac{1}{z_{i_1 j}} \tilde{g}_j \left( \frac{z_{i_1 j}}{z_{i_2 j}}, Z_{i_1 i_2, j} \right) \quad (65)$$

where $Z_{i_1 i_2, j}$ stands for

$$(z_{1j}, \ldots, z_{i_1-1, j}, z_{i_1+1, j}, \ldots, z_{i_2-1, j}, z_{i_2+1, j}, \ldots, z_{k-1, j})$$

and where $\tilde{g}_j$ is an arbitrary function having one argument less than $g_j$, or also by Lemma 3,

$$g_j(Z_j)_{i_1} - g_j(Z_j)_{i_2} = \frac{1}{z_{i_2 j}} \tilde{g}_j \left( \frac{z_{i_1 j}}{z_{i_2 j}}, Z_{i_1 i_2, j} \right) \quad (66)$$

where again $\tilde{g}_j$ is an arbitrary function having one argument less than $g_j$. Similarly, since $f_i$ and $g_j$ play a symmetric role in Eq. 9 as shown by Eq. 12 and hence have the same form, we get

$$f_i(W_i)_{j_1} - f_i(W_i)_{j_2} = \frac{x_i}{z_{ij_1} y_{j_1}} \tilde{f}_i \left( \frac{z_{ij_1} y_{j_1}}{z_{ij_2} y_{j_2}}, W_{j_1 j_2, i} \right) \quad (67)$$

where $W_{j_1 j_2, i}$ stands for

$$\left( \frac{z_{i1} y_1}{x_i}, \ldots, \frac{z_{i, j_1-1} y_{j_1-1}}{x_i}, \frac{z_{i, j_1+1} y_{j_1+1}}{x_i}, \ldots, \right.$$
$$\left. \frac{z_{i, j_2-1} y_{j_2-1}}{x_i}, \frac{z_{i, j_2+1} y_{j_2+1}}{x_i}, \ldots, \frac{z_{in} y_n}{x_i} \right)$$

or also, we have,

$$f_i(W_i)_{j_1} - f_i(W_i)_{j_2} = \frac{x_i}{z_{ij_2} y_{j_2}} \tilde{f}_i \left( \frac{z_{ij_1} y_{j_1}}{z_{ij_2} y_{j_2}}, W_{j_1 j_2, i} \right) \quad (68)$$

Now, by setting $j = j_1$ and $j = j_2$ in Eq. 54 and subtracting the resulting equations, we get,

$$\frac{g_{j_1}(Z_{j_1})_{i_1} - g_{j_1}(Z_{j_1})_{i_2}}{y_{j_1}} - \frac{g_{j_2}(Z_{j_2})_{i_1} - g_{j_2}(Z_{j_2})_{i_2}}{y_{j_2}} = \quad (69)$$
$$\frac{f_{i_1}(W_{i_1})_{j_1} - f_{i_1}(W_{i_1})_{j_2}}{x_{i_1}} - \frac{f_{i_2}(W_{i_2})_{j_1} - f_{i_2}(W_{i_2})_{j_2}}{x_{i_2}}$$

Plugging Eqs. 65 through 68 into Eq. 69 yields,

$$\frac{\tilde{g}_{j_1} \left( \frac{z_{i_1 j_1}}{z_{i_2 j_1}}, Z_{i_1 i_2, j_1} \right)}{z_{i_1 j_1} y_{j_1}} - \frac{\tilde{g}_{j_2} \left( \frac{z_{i_1 j_2}}{z_{i_2 j_2}}, Z_{i_1 i_2, j_2} \right)}{z_{i_2 j_2} y_{j_2}} = \quad (70)$$
$$\frac{\tilde{f}_{i_1} \left( \frac{z_{i_1 j_1} y_{j_1}}{z_{i_1 j_2} y_{j_2}}, W_{j_1 j_2, i_1} \right)}{z_{i_1 j_1} y_{j_1}} - \frac{\tilde{f}_{i_2} \left( \frac{z_{i_2 j_1} y_{j_1}}{z_{i_2 j_2} y_{j_2}}, W_{j_1 j_2, i_2} \right)}{z_{i_2 j_2} y_{j_2}}$$

Note that the variables in $Z_{i_1 i_2, j_1}$ do not appear elsewhere in this equation. Therefore, $\tilde{g}_{j_1}$ is only a function of its first argument. Similarly, $\tilde{g}_{j_2}$, $\tilde{f}_{i_1}$ and $\tilde{f}_{i_2}$ are only functions of their first argument. Thus Eq. 70 can be rewritten as follows,

$$\frac{1}{z_{i_1 j_1} y_{j_1}} \tilde{g}_{j_1} \left( \frac{z_{i_1 j_1}}{z_{i_2 j_1}} \right) - \frac{1}{z_{i_2 j_2} y_{j_2}} \tilde{g}_{j_2} \left( \frac{z_{i_1 j_2}}{z_{i_2 j_2}} \right) = \quad (71)$$
$$\frac{1}{z_{i_1 j_1} y_{j_1}} \tilde{f}_{i_1} \left( \frac{z_{i_1 j_1} y_{j_1}}{z_{i_1 j_2} y_{j_2}} \right) - \frac{1}{z_{i_2 j_2} y_{j_2}} \tilde{f}_{i_2} \left( \frac{z_{i_2 j_1} y_{j_1}}{z_{i_2 j_2} y_{j_2}} \right)$$

Let, $x = z_{i_1 j_1} y_{j_1}$, $y = z_{i_2 j_1} y_{j_1}$, $z = z_{i_1 j_2} y_{j_2}$, and $w = z_{i_2 j_2} y_{j_2}$ in Eq. 71. Then,

$$\frac{1}{x} \left[ \tilde{g}_{j_1} \left( \frac{x}{y} \right) - \tilde{f}_{i_1} \left( \frac{x}{z} \right) \right] = \frac{1}{w} \left[ \tilde{g}_{j_2} \left( \frac{z}{w} \right) - \tilde{f}_{i_2} \left( \frac{y}{w} \right) \right] \quad (72)$$

By taking a derivative wrt $y$ of Eq. 72, we get,

$$\frac{\tilde{g}'_{j_1}(\frac{x}{y})}{y^2} = \frac{\tilde{f}'_{i_2}(\frac{y}{w})}{w^2} \quad (73)$$

Setting $y = w$, we see that $\tilde{\tilde{g}}'_{j_1}(t) = \beta_{j_1}$ and $\tilde{\tilde{g}}_{j_1}(t) = \beta_{j_1} t + \alpha_{j_1}$ where $\alpha_{j_1}$ and $\beta_{j_1}$ are constants. Plugging this result into Eq. 65 yields,

$$g_j(Z_j)_{i_1} - g_j(Z_j)_{i_2} = \frac{\alpha}{z_{i_1 j}} + \frac{\beta}{z_{i_2 j}} \quad (74)$$

where $1 \leq i_1 < i_2 \leq k - 1$.

Eq. 74 is a first-order partial differential equation the general solution of which is given by Lemma 4. Consequently, due to Eq. 48, we get,

$$g_j(t_1, \ldots, t_{k-1}) = t_{i_1}^{\alpha_{i_1 j}} t_{i_2}^{\alpha_{i_2 j}} g_j(t_{i_1} + t_{i_2}, t_1, \ldots,$$
$$t_{i_1-1}, t_{i_1+1}, \ldots, t_{i_2-1}, t_{i_2+1}, \ldots, t_{k-1}) \quad (75)$$

Now due to Lemma 5, we have,

$$g_j(t_1, \ldots, t_{k-1}) = \left[\prod_{i=1}^{k-1} t_i^{\alpha_{ij}}\right] G_j\left(\sum_{i=1}^{k-1} t_i\right) \quad (76)$$

Similarly,

$$f_i(t_1, \ldots, t_{n-1}) = \left[\prod_{j=1}^{n-1} t_j^{\beta_{ij}}\right] F_i\left(\sum_{j=1}^{n-1} t_j\right) \quad (77)$$

which is obtained by repeating the derivation starting at Eq. 12 rather then at Eq. 9. Note that we have almost derived the Dirichlet functional form. It remains to derive the form of the functions $F_i$ and $G_j$.

In Eq. 9 let $z_{1j} = z_{2j} = \cdots = z_{kj}$ for $1 \leq j \leq n$. Thus, according to Eq. 10, $z_{ij} = x_i$. Consequently, we get,

$$f_0(y_1, \ldots, y_{n-1}) \prod_{j=1}^n g_j(x_1, \ldots, x_{k-1}) =$$
$$g_0(x_1, \ldots, x_{k-1}) \prod_{i=1}^k f_i(y_1, \ldots, y_{n-1}) \quad (78)$$

Eqs. 9 and 78 yield,

$$\prod_{j=1}^n \frac{g_j(z_{1,j}, \ldots, z_{k-1,n})}{g_j(x_1, \ldots, x_{k-1})} = \prod_{i=1}^k \frac{f_i\left(\frac{z_{i1}y_1}{x_i}, \ldots, \frac{z_{i,n-1}y_{n-1}}{x_i}\right)}{f_i(y_1, \ldots, y_{n-1})} \quad (79)$$

Plugging Eqs. 76 and 77 into Eq. 79, we get,

$$\left[\prod_{j=1}^n \prod_{i=1}^{k-1} \left(\frac{z_{ij}}{x_i}\right)^{\alpha_{ij}}\right] \left[\prod_{j=1}^n \frac{G_j\left(\sum_{i=1}^{k-1} z_{ij}\right)}{G_j\left(\sum_{i=1}^{k-1} x_i\right)}\right] =$$
$$\left[\prod_{i=1}^k \prod_{j=1}^{n-1} \left(\frac{z_{ij}}{x_i}\right)^{\beta_{ij}}\right] \left[\prod_{i=1}^k \frac{F_i\left(\sum_{j=1}^{n-1} \frac{z_{ij}y_j}{x_i}\right)}{F_i\left(\sum_{j=1}^{n-1} y_j\right)}\right] \quad (80)$$

Thus, using $z_{kj} = 1 - \sum_{i=1}^{k-1} z_{ij}$ (Eq. 10),

$$\left[\prod_{j=1}^{n-1} \prod_{i=1}^{k-1} \left(\frac{z_{ij}}{x_i}\right)^{c_{ij}}\right] \left[\prod_{j=1}^n \frac{\tilde{G}_j\left(\sum_{i=1}^{k-1} z_{ij}\right)}{\tilde{G}_j\left(\sum_{i=1}^{k-1} x_i\right)}\right] =$$
$$\prod_{i=1}^k \frac{\tilde{F}_i\left(\sum_{j=1}^{n-1} \frac{z_{ij}y_j}{x_i}\right)}{\tilde{F}_i\left(\sum_{j=1}^{n-1} y_j\right)} \quad (81)$$

where for $1 \leq i \leq k-1$ and $1 \leq j \leq n-1$, $c_{ij} = \alpha_{ij} - \beta_{ij}$,

$$\tilde{F}_i(t) = (1-t)^{-\alpha_{in}} F_i(t) \quad \tilde{G}_j(t) = (1-t)^{-\beta_{kj}} G_j(t)$$

and where $\tilde{F}_k(t) = F_k(t)$ and $\tilde{G}_n(t) = G_n(t)$. We will show that $\tilde{F}_i(t)$, $i = 1, \ldots, k-1$, are constants. Consequently, due to Eq. 77, $f_i$ has a Dirichlet functional form. That the function $f_k$ also has a Dirichlet functional form can be obtained by choosing $z_{1j}$ as a dependent variable defined by $z_{1j} = 1 - \sum_{i=2}^k z_{ij}$ instead of $z_{kj}$ as defined by Eq. 10 and repeating the same arguments. By symmetric arguments each $g_j$ also has a Dirichlet functional form.

Let $y_j = \frac{1}{n}$, for all $j$, $1 \leq j \leq n$ and $z_{ij} = \frac{1}{k}$ for all $i$ and $j$, $1 \leq i \leq k$, $1 \leq j \leq n-1$. Hence, the only free variables remaining in Eq. 81 are $z_{in}$ where $1 \leq i \leq k-1$. Note that $x_i = \sum_{j=1}^n z_{ij} y_j = \frac{n-1}{kn} + \frac{1}{n} z_{in}$, $1 \leq i \leq k-1$, and so $\tilde{G}_j(\sum_{i=1}^{k-1} x_i)$ is a function of $\sum_{i=1}^{k-1} z_{in}$. Also $\tilde{G}_j(\sum_{i=1}^{k-1} z_{ij})$ is a constant for $1 \leq j \leq n-1$ and a function of $\sum_{i=1}^{k-1} z_{in}$ for $j = n$. Consequently, Eq. 81 becomes,

$$f\left(\sum_{i=1}^{k-1} z_{in}\right) = \prod_{i=1}^{k-1} \tilde{F}_i\left(\frac{c}{c + d z_{in}}\right) \left[\frac{c + d z_{in}}{c}\right]^{a_i} \quad (82)$$

where $c = \frac{n-1}{kn}$, $d = \frac{1}{n}$ and $a_i = \sum_{j=1}^{n-1} c_{ij}$. Note that $z_{kn} = 1 - \sum_{i=1}^{k-1} z_{in}$ and so the k-th term on the right hand side of Eq. 81 is absorbed, along with some constants, into the definition of $f$ in Eq. 82.

Let $t_i = \frac{c}{c + d z_{in}}$; $z_{in} = \frac{c}{d} \frac{1-t_i}{t_i}$. Taking the logarithm of Eq. 82, we get,

$$\hat{f}\left(\frac{c}{d} \sum_{i=1}^{k-1} \frac{1-t_i}{t_i}\right) = \sum_{i=1}^{k-1} \ln t_i^{-a_i} \tilde{F}_i(t_i) \quad (83)$$

Taking a derivative wrt $t_{i_1}$, $1 \leq i_1 \leq k-1$, we get,

$$-\frac{c}{d t_{i_1}^2} \hat{f}'\left(\frac{c}{d} \sum_{i=1}^{k-1} \frac{1-t_i}{t_i}\right) = \left[\ln t_{i_1}^{-a_{i_1}} \tilde{F}_{i_1}(t_{i_1})\right]' \quad (84)$$

Thus, $\hat{f}'\left(\frac{c}{d} \sum_{i=1}^{k-1} \frac{1-t_i}{t_i}\right)$ must be a constant. Hence, by integrating Eq. 84,

$$\tilde{F}_i(t) = c_i t^{a_i} e^{\frac{K}{t}}, \quad 1 \leq i \leq k-1 \quad (85)$$

where $K$ is a constant not depending on $i$.

To complete the derivation we substitute Eq. 85 into Eq. 81, let $y_j = \frac{1}{n}$, for $1 \leq j \leq n$ and $z_{ij} = \frac{1}{k}$ except $z_{i1}$, $1 \leq i \leq k-1$ which remain free variables. Consequently, we get

$$g\left(\sum_{i=1}^{k-1} z_{i1}\right) = \prod_{i=1}^{k-1} \frac{(z_{i1} + w_0)^{a_i}}{z_{i1}^{c_{i1}}} e^{K \sum_{i=1}^{k-1} \frac{1}{z_{i1} + w_0}}$$

where $w_0 = \frac{n-2}{k}$. Therefore, $K = 0$, $a_i = 0$, and $\tilde{F}_i$ is a constant as claimed.

Thus,

$$f_i(t_1,\ldots,t_{n-1}) = k_i \left[\prod_{j=1}^{n-1} t_j^{\beta_j}\right] (1 - \sum_{j=1}^{n-1} t_j)^{\beta_k} \quad (86)$$

$$g_j(t_1,\ldots,t_{k-1}) = c_j \left[\prod_{i=1}^{k-1} t_i^{\alpha_i}\right] (1 - \sum_{i=1}^{k-1} t_i)^{\alpha_k} \quad (87)$$

We now comment on how the derivation changes when $n = 2$ and $k \geq 3$. The case $n \geq 3$ and $k = 2$ follows as well due to the symmetric functional equation (Equation 12).

Note that up to Eq. 57 the derivation is valid when $n = 2$. Furthermore, note that the sum in Eq. 56 consists now of one term where $l = j_1 = 1$. Thus, Eq. 56 and Eq. 57 yield, using $x_i = z_{ij_1} y_{j_1} + z_{in} y_n$ (n=2, $j_1 = 1$),

$$\frac{f_{i_1}(W_{i_1})_{j_1}}{x_{i_1}} = \frac{z_{i_1 n}}{y_n}[g_n(Z_n)_{i_1 i_1} - g_n(Z_n)_{i_2 i_1}] -$$
$$\frac{z_{i_1 j_1}}{y_{j_1}}[g_{j_1}(Z_{j_1})_{i_1 i_1} - g_{j_1}(Z_{j_1})_{i_2 i_1}] \quad (88)$$

Similarly,

$$\frac{f_{i_2}(W_{i_2})_{j_1}}{x_{i_2}} = \frac{z_{i_2 n}}{y_n}[g_n(Z_n)_{i_1 i_2} - g_n(Z_n)_{i_2 i_2}] -$$
$$\frac{z_{i_2 j_1}}{y_{j_1}}[g_{j_1}(Z_{j_1})_{i_1 i_2} - g_{j_1}(Z_{j_1})_{i_2 i_2}] \quad (89)$$

which is obtained by taking a derivative wrt $z_{i_2 j_1}$ of Eq. 54 (instead of wrt $z_{i_1 j_1}$) and repeating the derivation up to Eq. 57.

Plugging Eqs. 88 and 89 into Eq. 54 and collecting all terms involving $y_n$ in one side and all terms not involving $y_n$ on the other side implies that each side is equal to a constant, say $c$, namely, we obtain the partial differential equation for $g_j(Z_j)$, $1 \leq j \leq n$, given by Eq. 62. Consequently, as given by Eq. 65 and because $n = 2$,

$$g_{j_1}(Z_{j_1})_{i_1} - g_{j_1}(Z_{j_1})_{i_2} = \frac{1}{z_{i_1 j_1}} \hat{g}_{j_1}(\frac{z_{i_1 j_1}}{z_{i_2 j_1}}) \quad (90)$$

and,

$$g_{j_2}(Z_{j_2})_{i_1} - g_{j_2}(Z_{j_2})_{i_2} = \frac{1}{z_{i_2 j_2}} \hat{g}_{j_2}(\frac{z_{i_1 j_2}}{z_{i_2 j_2}}) \quad (91)$$

Also, when $n = 2$, we have $x_i = z_{ij_1} y_{j_1} + z_{in} y_n$, and hence,

$$\frac{1}{x_{i_1}} f_{i_1}(W_{i_1})_{j_1} = \frac{1}{x_{i_1}} f_{i_1}(\frac{z_{i_1 j_1} y_{j_1}}{x_{i_1}})_{j_1}$$
$$= \frac{1}{z_{i_1 j_1} y_{j_1}} \hat{f}_{i_1}(\frac{z_{i_1 j_1} y_{j_1}}{z_{i_1 n} y_n}) \quad (92)$$

$$\frac{1}{x_{i_2}} f_{i_2}(W_{i_2})_{j_1} = \frac{1}{x_{i_2}} f_{i_2}(\frac{z_{i_2 j_1} y_{j_1}}{x_{i_2}})_{j_1}$$
$$= \frac{1}{z_{i_2 j_1} y_{j_1}} \hat{f}_{i_2}(\frac{z_{i_2 j_1} y_{j_1}}{z_{i_2 n} y_n}) \quad (93)$$

Plugging Eqs. 90 and 93 into Eq. 54 yields,

$$\frac{1}{z_{i_1 j_1} y_{j_1}} \hat{g}_{j_1}(\frac{z_{i_1 j_1}}{z_{i_2 j_1}}) - \frac{1}{z_{i_2 n} y_n} \hat{g}_n(\frac{z_{i_1 n}}{z_{i_2 n}}) = \quad (94)$$
$$\frac{1}{z_{i_1 j_1} y_{j_1}} \hat{f}_{i_1}(\frac{z_{i_1 j_1} y_{j_1}}{z_{i_1 n} y_n}) - \frac{1}{z_{i_2 n} y_n} \hat{f}_{i_2}(\frac{z_{i_2 j_1} y_{j_1}}{z_{i_2 n} y_n})$$

This equation parallels Eq. 71 where ($j_2$ is replaced by $n$) and can be solved in the same way. Thus Eq. 76 is obtained. Eq. 77, on the other hand, needs no proof when $n = 2$ because an arbitrary function $f(x)$ defined on $(0,1)$ can always be written as $f(x) = x^\alpha g(x)$ where $g(x) = x^{-\alpha} f(x)$. The rest of the derivation follows closely the previous section.

## The Joint Distribution

We have so far shown that, under the assumptions made by Theorem 2, $f_I(\theta_{I\cdot})$ and $f_{J|i}(\theta_{J|i})$ are Dirichlet. Similarly, $f_J(\theta_{\cdot J})$ and $f_{I|j}(\theta_{I|j})$ have been shown to be Dirichlet as well. We now show that if $f_I$, $f_{J|i}$, $f_J$ and $f_{I|j}$ are all Dirichlet, then the joint distribution $f_U(\{\theta_{ij}\})$ must also be a Dirichlet.

We can write,

$$f_{JI}(\theta_{\cdot J}, \theta_{I|1}, \ldots, \theta_{I|k}) = f_J(\theta_{\cdot J}) \prod_{j=1}^n f_{I|j}(\theta_{I|j})$$
$$= c \prod_{j=1}^k \theta_{\cdot j}^{\alpha_j - 1} \prod_{j=1}^k \prod_{i=1}^n \theta_{i|j}^{\alpha_{i|j} - 1}$$

But $f_{IJ}(\theta_{I\cdot}, \theta_{J|1}, \ldots, \theta_{J|k})$ can be expressed using $f_{JI}$ by two applications of the Jacobian given by Eq. 3. Thus we get,

$$f_{IJ}(\theta_{I\cdot}, \theta_{J|1}, \ldots, \theta_{J|k}) \propto \left[\prod_{i=1}^k \theta_{i\cdot}^{n-1}\right] \left[\prod_{j=1}^n \theta_{\cdot j}^{k-1}\right]^{-1}$$
$$\left[\prod_{j=1}^k \theta_{\cdot j}^{\alpha_j - 1} \prod_{j=1}^k \prod_{i=1}^n \left[\frac{\theta_{j|i} \theta_{i\cdot}}{\theta_{\cdot j}}\right]^{\alpha_{i|j} - 1}\right] \quad (95)$$

where $\theta_{\cdot j} = \sum_i \theta_{i\cdot} \theta_{j|i}$. Since $f_{IJ}$ can be expressed, due to local and global independence, as a product of $f_I$, $f_{J|1}, \ldots, f_{I|n}$ each of which has been shown to be a Dirichlet, it follows from Eq. 95 that the exponent coefficients for $\theta_{\cdot j}$, $1 \leq j \leq n$, must vanish. Consequently, $f_U(\{\theta_{ij}\})$, which is obtained from Eq. 95 by multiplying with $\{\prod_{i=1}^k \theta_{i\cdot}^{n-1}\}^{-1}$ and using the relationship $\theta_{ij} = \theta_{j|i} \theta_{i\cdot}$, is Dirichlet.